# A Novel Scheme for Intelligent Recognition of Pornographic Images

Seyed Mostafa Kia[1,2], Hossein Rahmani[3], Reza Mortezaei[4], Mohsen Ebrahimi Moghaddam[4], Amer Namazi[5]

[1] *Neuroinformatics Laboratory (NILab), Bruno Kessler Foundation, Trento, Italy*
[2] *University of Trento, Trento, Italy*
[3] *University of Western Australia, Crawley, Australia*
[4] *Electrical and Computer Engineering Department, Shahid Beheshti University; G.C; Tehran, Iran*
[5] *Douran Software Technologies Co., Tehran, Iran*

***Abstract.*** *Harmful contents are rising in internet day by day and this motivates the essence of more research in fast and reliable obscene and immoral material filtering. Pornographic image recognition is an important component in each filtering system. In this paper, a new approach for detecting pornographic images is introduced. In this approach, two new features are suggested. These two features in combination with other simple traditional features provide decent difference between porn and non-porn images. In addition, we applied fuzzy integral based information fusion to combine MLP (Multi-Layer Perceptron) and NF (Neuro-Fuzzy) outputs. To test the proposed method, performance of system was evaluated over 18354 download images from internet. The attained precision was 93% in TP and 8% in FP on training dataset, and 87% and 5.5% on test dataset. Achieved results verify the performance of proposed system versus other related works.*

***Keywords:*** *MLP, NF,* Pornography detection, Skin detection,

## I. Introduction

One of the main concerns in using the Internet is the existence of harmful (e.g. pornographic) contents and these non-desired materials are growing at increasing rate. According to *[1]*, 25% of the queries in search engines, 8% of emails, and 12% of homepages are porn-related. Therefore, recognizing harmful images is a big challenge in filtering systems.

Filtering methods for Pornographic contents are categorized in three major classes: (1) filtering the keywords for text contents related to pornographic contents, (2) filtering the access to these prohibited sites by collecting lists of adult website addresses, and (3) content-based analysis of images for detecting offensive materials. This last alternative has the benefit of allowing an adaptive, text-independent, and dynamic purifying of pornographic contents.

The first effort on classifying nude pictures has been done by Forsyth et al. *[2,3,4]*. In this technique, the images with large skin regions are detected and then using a particularly defined human structure *[5]*, the images containing human body parts are recognized as the nude ones. In this method, only the images with skin area larger than one third are fed to a geometric analyzer. The main drawback of this method is time inefficiency. Wang et al. developed a system named WIPE which employs Daubechies wavelets, normalized central moments, and color histograms to provide semantically-meaningful feature vector matching *[6]*. Jones and Rehg detect skin color by approximating the distribution of skin and non-skin color in the color space. After that, some simple features are extracted to detect pornographic images *[7]*. Bosson et al. proposed a pornography detection system which is also based on skin detection *[8]*. Zheng et al. construct a maximum entropy model for skin detection, and then nine low level geometric features are extracted from the detected skin region. Finally a Multi-Layer Perceptron (MLP) classifier is trained for image classification by using these features *[9]*. In *[10]*, adult image detection scheme is suggested by using the skin color in the YUV, YIQ, and their polar representations (YUV°, YIQ°). In this approach, a Sobel edge detector and a Gabor filter verify the results of the skin detector. In *[11]*, the researchers used the CIE-Lab color space for detecting the skin areas in the images. The results were investigated using some geometrical features.

Skin detection results can significantly affect the subsequent feature extraction and pornographic classification. Different researchers have used different techniques such as explicit skin-color space thresholding, histogram model with Bayes classification, Gaussian classifier and etc. in different color space. But, each of these techniques has its own pros and cons *[12]*.

In this paper, we propose a new method to detect pornography images with high precision. In this method, several features which are simple and time efficient, are extracted from skin region and whole image. For the first time, the Fourier descriptors and signature of boundary of skin region is used as shape descriptor features. Next, important features for pornography which are more distinctive and less correlated are found based on Self Organizing Feature Maps (SOFM) and correlation analysis. Finally, the selected features are fed to parallel classifiers and the output of each classifier is sent to fuzzy decision making component. To assess the performance of proposed method, we compared this system with three other methods on the collected dataset.

The rest of paper is organized as follows: In section 2, the proposed method is introduced. Section 3 describes the experimental results and in section 4 the paper is concluded.

## II. PROPOSED APPROACH

In this section, the proposed scheme for pornographic image detection is introduced. The Figure 1 illustrates the architecture of proposed system.

In this system after some preprocessing such as down-sampling, the histogram model provided by [7] is used for skin detection. At that point, the feature vector is extracted from largest skin region. Then, *2* different classifies are fed with the same extracted feature vectors in parallel manner. Finally the outputs of classifiers are combined to make the final decision.

In the following subsections, the feature extraction and selection procedures are discussed in details. Then, the suggested classification system will be described in more details.

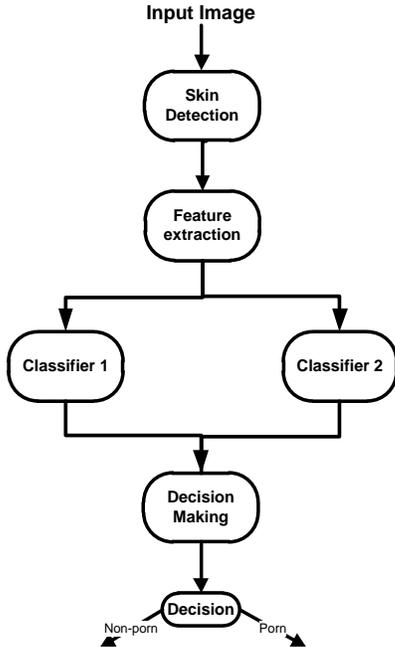

**Figure 1 - The Architecture of proposed System**

### A. Feature Extraction

As mentioned before, in this study, we employ the histogram model for skin detection. Since this skin detection method is pixel based, some non-skin pixels with colors similar to skin may be detected as skin pixels. Therefore, the morphological operations "opening and closing" are applied on binary output (0 for non-skin and 1 for skin pixels) of skin detection to refine the skin regions. For each morphological operation, a disk shape structural element is used. The radius of the disk is set automatically according to the size of the image by Eq. 1.

$$Disk\ Radial = \frac{(x+y)}{c} \quad (1)$$

Where *x* and *y* are length and width of image and *c* is a constant number. The value for c was set 75 for opening operation and 100 for closing operation by experiment.

In the literature, different features have been used to classify the non-porn from porn images [9,13,14,15]. These features can be categorized into three main types: 1) quantitative features related to structure of largest skin region, 2) shape descriptor features which contain information about shape of skin regions, and 3) features related to texture of detected skin regions. In this paper, two new features are presented that improve the performance of classifiers in discriminating between portrait and porno images: Fourier descriptors and number of peaks in signature of largest region boundary.

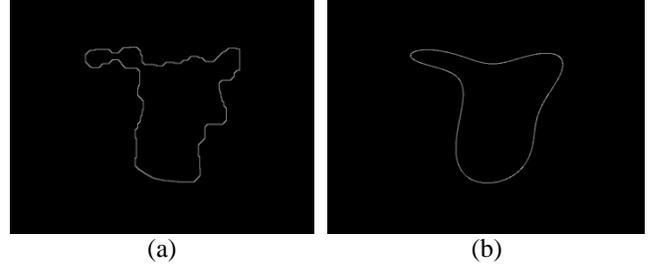

(a) (b)

**Figure 2 – (a) largest region boundary, (b) reconstructed boundary using 10 descriptors**

*1) Fourier descriptors*

Fourier descriptors describe the shape in terms of its spatial frequency content. Fourier descriptors are the boundary based descriptors in 2-D space using the Fourier methods [16]. The K-point digital boundary in the xy-plane starting at an arbitrary point $(x_0, y_0)$, coordinate pairs $(x_0, y_0), (x_1, y_1) \ldots (x_{k-1}, y_{k-1})$ are met in pass through the boundary, in counterclockwise. With this definition, the boundary itself can be represented as sequence of coordinates $s(k) = [x(k), y(k)]$ (k = 0, 1, 2, ..., K-1). Each coordinate pair can be treated as $s(k) = x(k) + jy(k)$. In this complex representation, x-axis is treated as real axis and the y-axis as the imaginary axis. The discrete Fourier transform (DFT) of s(k) is :

$$a(u) = \sum_{k=0}^{K-1} s(k) e^{-j2\pi u k / K} \quad (2)$$

The complex coefficients $a(u)$ are called the Fourier descriptors of the boundary. The inverse Fourier transform of these coefficients restores s(k) (Eq. 3).

$$s(k) = \frac{1}{K} \sum_{u=0}^{K-1} a(u) e^{j2\pi u k / K} \quad (3)$$

The *0*-th Fourier descriptor has no information about shape of boundary. This descriptor contains information about position of largest region but other descriptors are translation invariant. The remained Fourier descriptors contain information about frequency of change in x-y coordination of boundary. So, these descriptor helps classifier to discriminate between smooth boundaries (usually face and rounded objects) and rough boundaries (like a body shape). This fact leads to superiority of our method because it covers necessity for face detection process. Therefore, time and computational complexity will reduce significantly. It must be mentioned that to make Fourier coefficients size invariant, all Fourier descriptors are divided by the magnitude of the 1st Fourier descriptor. Furthermore, to provide orientation and starting point invariance, only the magnitude of the Fourier descriptors is considered. In this study, 10 Fourier descriptors (0th and 2nd-11th) are included in the final feature vector. Figure 2 shows reconstruction of instance boundary using these 10 coefficients and inverse Fourier transform. As obvious, these 10 descriptors contain information about overall shape of boundary. Figure 3 illustrates the different distribution of the $0^{th}$ and the $10^{th}$ Fourier descriptors over number of porn and non-porn images. Therefore, Fourier descriptors can be stated as good judicious factor between two target classes.

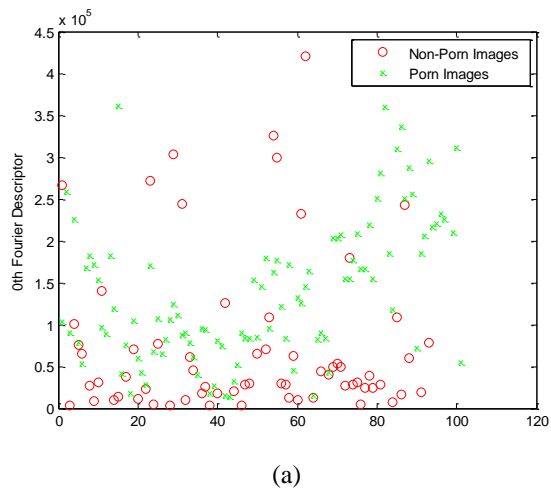

(a)

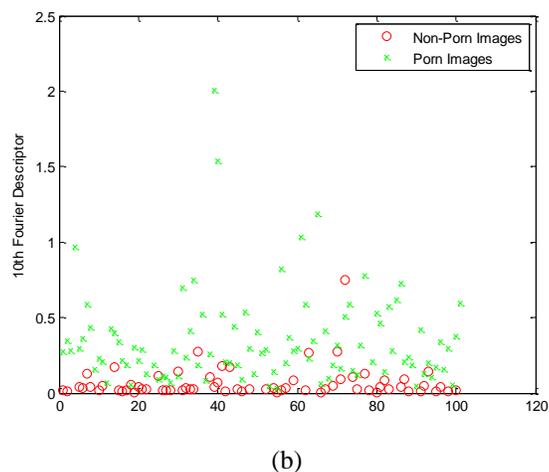

(b)

**Figure 3- Distribution of (a) $0^{th}$ and (b) $10^{th}$ Fourier descriptors over 200 samples of porn and non-porn images**

*2) Boundary Signature*

As mentioned before, the second new feature of our work is the number of peaks in signature of largest skin region. In general, a signature is any 1-D function which representing 2-D areas or boundaries. One of the common methods to provide signature of a boundary is to plot the distance from an interior point (e.g., the centroid) to the boundary as a function of angle. Figure 4 shows examples of signatures for a circle and square [17].

In this way, number of peaks in signature of boundary has some tangible properties about its overall shape. To show competence, Figure 5 illustrates different concentration in arrangement of this feature for 1000 porn and non-porn images. Despite overlap between two categories about some samples, apparently the number of peaks in porn images is more than non-Porn images.

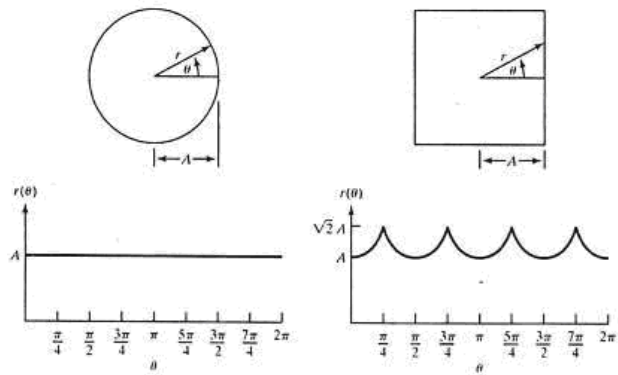

**Figure 4- Examples for Signature of boundary for a circle and a square (reprinted from [17])**

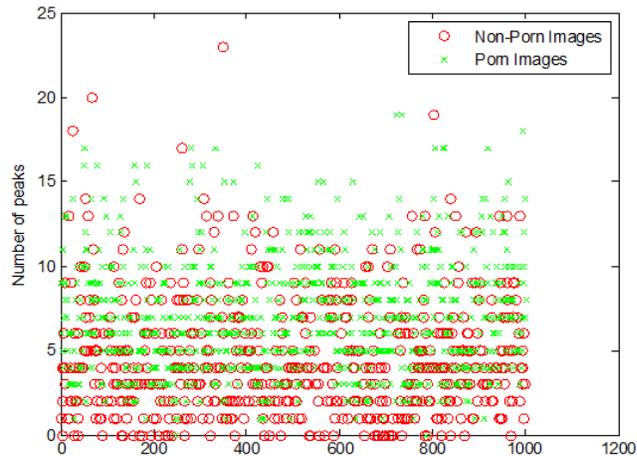

**Figure 5- Distribution of peak number of signature over number of porn and non-porn images**

*B. Feature Selection*

After extracting new suggested features together with features which are introduced by others, a vector of 108 features is provided for each image. Then, the SOFM Neural Network is used for visualizing clusters [18]. This process alongside correlation analysis has been applied to eliminate the correlated features [19]. Figure 6 shows an instance of analysis has been done by weight planes of SOFM Neural Network to eliminate redundant features. For example, similarity of weight planes about input 3 and 5 demonstrates high correlation between them. Furthermore, uniformity of weight plane of input 3 comparing to other weight planes shows less effectiveness in clustering process.

In this way, the number of elements in feature vector (selected features) decreased to 38. The selected features have been listed as follows:
1- The ratio of weighted skin pixels to image size

2-Number of connected components
3-10 Fourier descriptors of boundary
4-Number of peaks on boundary signature of largest region
5-Eccentricity of largest region
6-Diameter of equivalent circle
7-The ratio of perimeter to area of largest region
8-Number of colors in image
9-First and second invariant moments of image
10-Center of global and local ellipse
11-Major and minor axis of global and local ellipse
12-The ratio of minor axis to major axis of global and local ellipse
13-Orientation difference between global and local ellipse
14-Solidity and extension of largest region
15-The ratio of width of bounding box to height of bounding box
16-The ratio of largest skin region to all skin regions
17-The edge direction histogram of largest skin region in 6 directions (i.e. 0-45-90-135-225-315°)

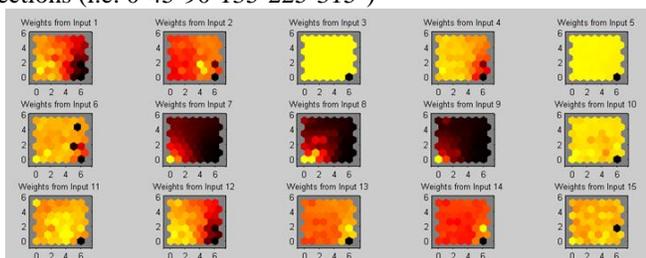

**Figure 6 – SOFM weight planes about 15 features of feature vector, darker colors represent larger weights between each input and SOFM neurons**

*C. Pornographic Classification*

Recently, the concept of combining multiple classifiers has been considered for developing highly accurate "diagnostic" systems [20]. Due to this fact, multi parallel classifier architecture is proposed to enhance the performance of pornography detection system. In the proposed approach, two parallel classifiers are exploited to classify the input images to two target categories. To find suitable classifiers, performance of MLP, SOFM, and Neuro-Fuzzy (NF) approach were investigated [21,22]. The configurations used for three mentioned classifiers are shown in Table 1.

**Table 1 - Configuration of different classifiers**

| Type of Classifier | Parameters |
|---|---|
| MLP | 38 input nodes, one hidden layer with 10 neurons, and 1 output neurons. Scaled conjugate gradient was chosen as the training algorithm. Also, the Cross-validation method used to validate the performance. |
| NF | Combination of subtracting fuzzy clustering and ANFIS, Cross-validation method used to validate the model. |
| SOFM | 38 input nodes and 16 neurons arranged in 2-D Hexagonal topology. |

After comparing performance of each of these classifiers, finally the MLP is chosen as first classifier and Neuro-Fuzzy (NF) as second. The details of implementation and comparison are described in experimental result section.

One of the key issues of this approach is how to combine the results of the various classifiers to provide the best approximation of the optimal result. Here, a fusion technique based on fuzzy integral is used. In the proposed method, fuzzy integral based information fusion proposed by [23] is used. For two inputs $h_1$ and $h_2$ the combination is:

$$M_{FI}(\mu, h) = [\mu(1,2) - (\mu(2) + \mu(1))]h_1 + \mu(1)h_1 + \mu(2)h_2 \qquad (4)$$

Where, µ is weight of importance. Figure 7 illustrates the overall diagram of system processes.

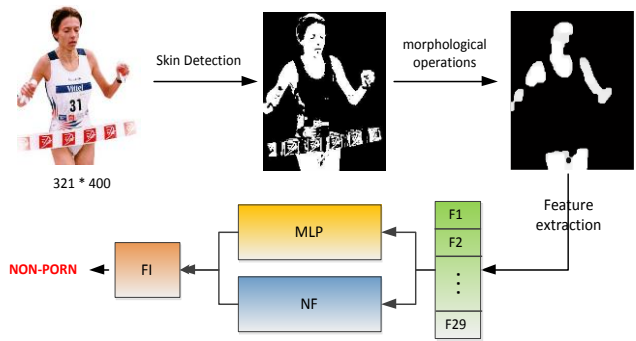

**Figure 7- Outline of system processes**

III. EXPERIMENTAL RESULTS

To assess the proposed method, a dataset with 18354 images (9295 pornography images and 9059 non-porn images) was prepared[1]. These images were downloaded from Internet by crawler and torrent. Also, the non-Porn images typically contain objects with colors very similar to human skin, such as portrait, body parts, athletes, desert, yellow flowers, lions, and so on. From these collected images, 8000 images (containing 4000 pornography images) were used for training the classifiers and 10354 images (containing 4295 pornography images) were used for testing the performance of proposed method. The development environment tool was MATLAB and the computation platform was a personal computer with 2.5GHz of CPU and 3 GB of RAM.

Two criteria used for performance evaluation are true positive (TP) and false positive (FP). TP shows the ratio of porn image which are truly recognized. Also FP represents the ratio of non-porn image which misclassified as porn images. The used configurations of classifiers have been mentioned in Table 1. To avoid over fitting problem in training process, the cross validation method is used to validate the training procedure. The training performance of MLP and NF are shown in Figure 8 and **Figure 9**. The values of TP and FP of each classifier are shown in Table 2. As shown in the table,

---
[1] . The dataset is available at:
https://drive.google.com/folderview?id=0B3ObURcsUUExZUVac29MRXRFbHM&usp=drive_web

while the MLP classifier reaches the highest TP, NF provides the best FP among the others.

As mentioned before, two classifiers based on proposed architecture (Figure 1) should be chosen. To select the two best classifiers, different combination of three classifiers like MLP+SOFM, MLP+NF, SOFM+NF have been tested and only the results of best pair (MLP+NF) are shown in Table 3. Table 3 also compares the proposed method with some related works that were implemented.

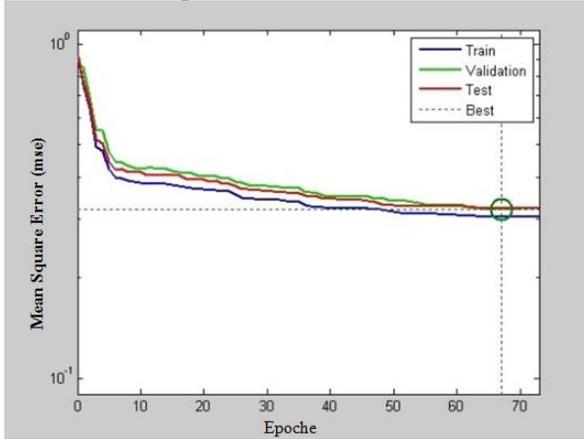

**Figure 8- Training curve for proposed MLP Neural Network**

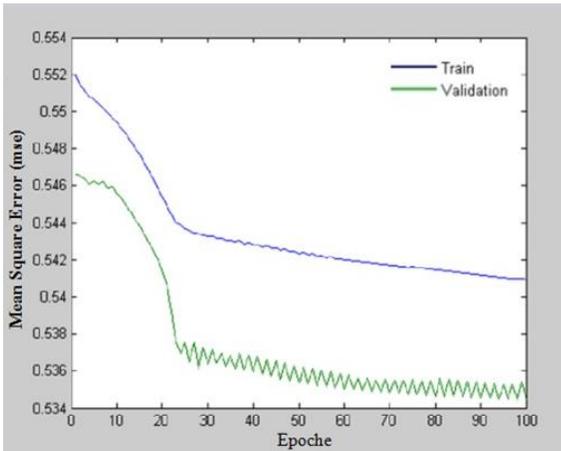

**Figure 9- Training curve for proposed Fuzzy Neural Network**

**Table 2- The values of TP and FP of different classifiers**

|    | MLP    | F-NN   | SOFM   |
|----|--------|--------|--------|
| TP | 87.21% | 84.48% | 82.97% |
| FP | 6.68%  | 5.35%  | 9.33%  |

To determine the values of μ in Eq. 4, system performance measured over training set by changing these values between 0 and 1 (Figure 10). In Figure 10, point A shows the best performance of combined classifier over training set. In addition, point B represents system performance ignoring MLP output and point C denotes system performance ignoring NF output. As it is obvious, performance of this parallel classifier is better than each single ones. Accordingly the values $\mu(h_1) = 0.47$, $\mu(h_2) = 0.53$ are attained for $h_1$ and $h_2$.

Figure 11 shows the classification results on some typical image samples. The misclassification of images (c) and (d) is due to similarity of objects in the images to the specific objects in the porn images. Other images are classified correctly.

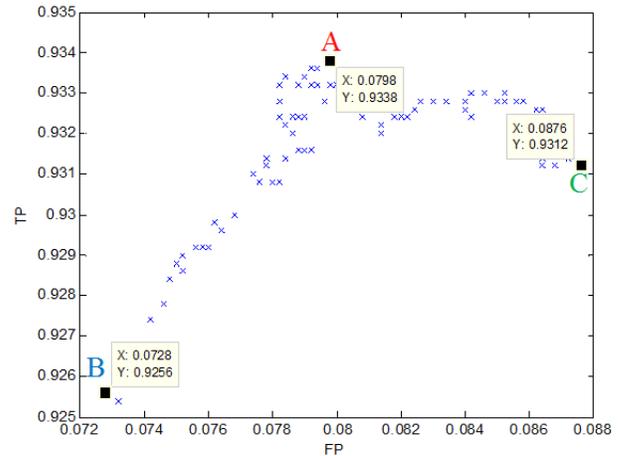

**Figure 10- Determining optimum parameters for FI**

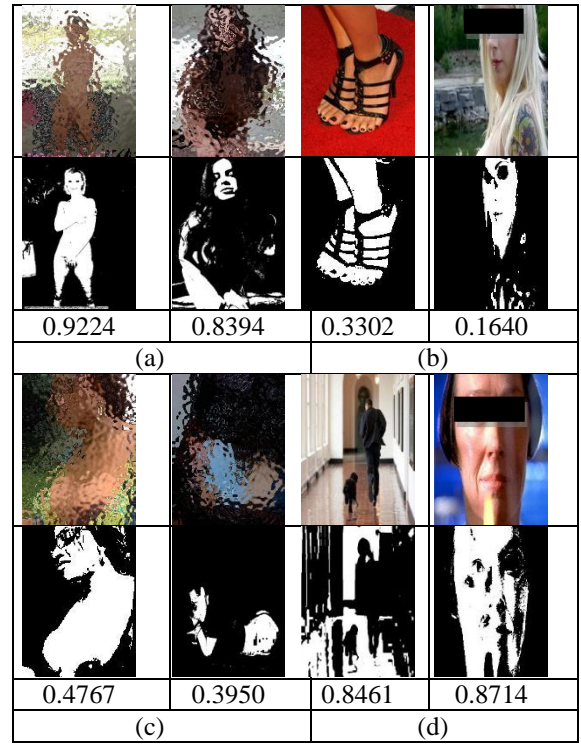

**Figure 11- Examples of image classification in the proposed system. The numbers show the output of proposed classification system which is positive (porn) when larger than 0.5 and negative (non-porn) otherwise. The images labeled as (a), (b), (c) and (d) are TP, TN (True Negative), FN (False Negative) and FP cases, respectively.**

## IV. CONCLUSION

In this paper, besides introducing new features, a novel scheme for classifying pornography images has been proposed. In this approach, a vector of effective features extracted from skin regions is used for training fuzzy integral based combination of two classifiers (MLP and NF). The proposed approach in comparison with other related works shows acceptable result in terms of TP and FP.

Although the experimental results are convincing but the results can be improved by extracting high level features by detecting objects such as face, breast and so on. Also more researches on type and number of classifiers in proposed architecture will lead to more accurate classifier.

**Table 3- Comparison between proposed method and other related methods**

|    | Proposed method (MLP+F-NN) | [13] | [15] | [14] |
|----|---|---|---|---|
| TP | 86.97% | 83.65% | 78.37% | 90.93% |
| FP | 5.38% | 6.64% | 16.98% | 18.94% |